\documentclass[10pt]{article}
\usepackage{amsfonts}
\usepackage{epsfig}
\usepackage{graphicx}
\usepackage{color}
\usepackage{titlepic}

\begin{document}




\title{\bf\huge{Automated languages phylogeny from Levenshtein distance} \\
\large{(Filogenia automatizada de l\'inguas
a partir da dist\^ancia de Levenshtein)}}

\author{Maurizio Serva
}
\date{}
\maketitle

{\centerline {\it Dipartimento di Matematica, Universit\`{a} 
dell'Aquila, I-67010 L'Aquila, Italy}} 
\bigskip
\bigskip

\bigskip
\bigskip

\includegraphics[scale=0.5]{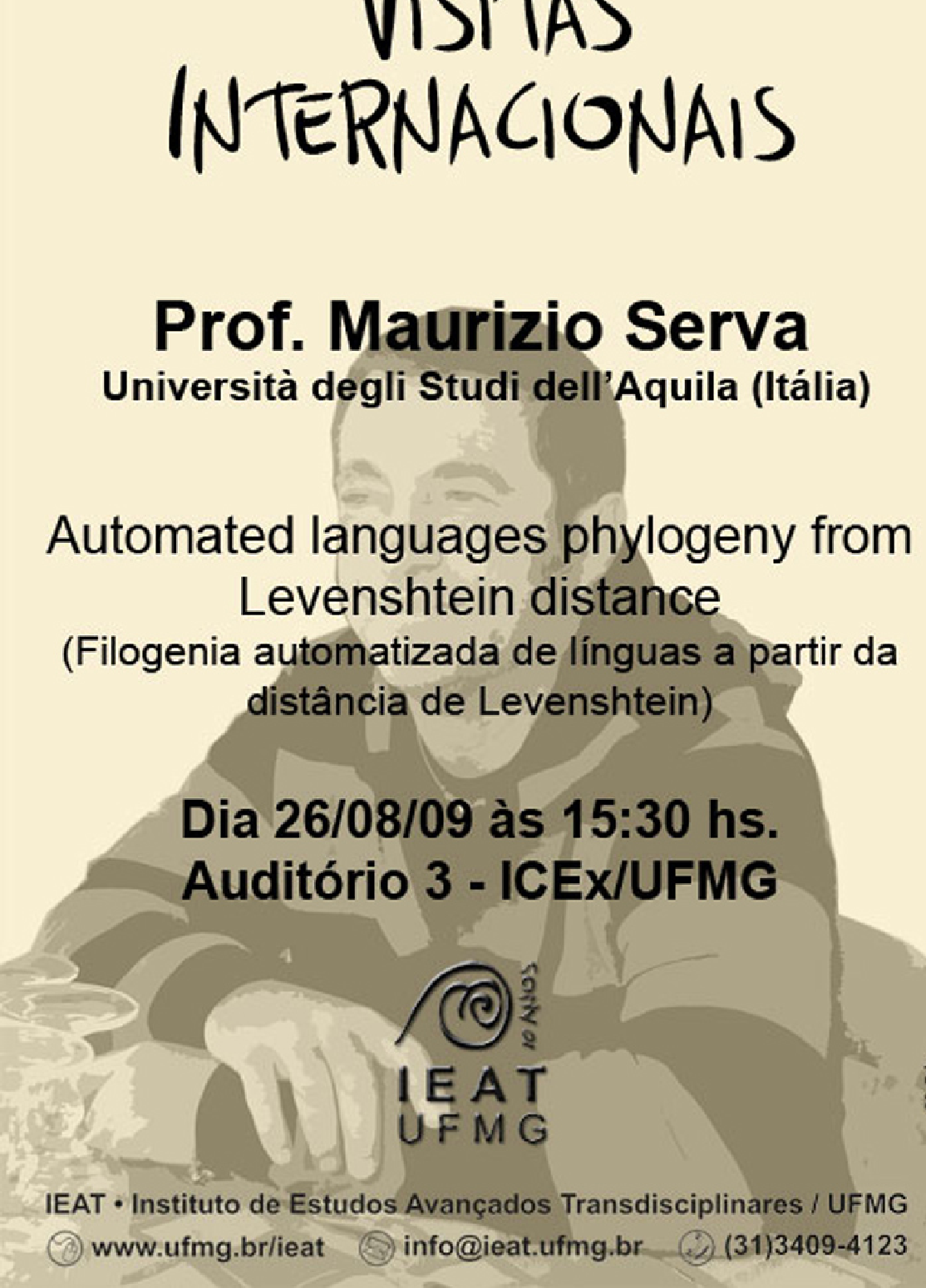}

\renewcommand{\thefootnote}{} 
\bigskip
\footnote{This paper contains the text of the invited talk given at the 
Universidade Federal de Minas Gerais:
{\it International Conference: Visitas Internationais, 
Instituto de Estudos Avan\c{c}ados Transdisciplinares}, 
(Belo Horizonte, 26 August 2009). 
https://www.ufmg.br/ieat/visitas-internacionais/}

\newpage

\begin{abstract}

\bigskip
\bigskip

L\'inguas evoluem com o tempo em um processo em que reprodu\c{c}\~{a}o, 
muta\c{c}\~{a}o e extin\c{c}\~{a}o 
s\~{a}o todos poss\'iveis, de forma semelhante ao que acontece 
com os organismos vivos. 
Usando esta similaridade \'e poss\'ivel, em princ\'ipio, 
construir \'arvores geneal\'ogicas que 
mostrem o grau de parentesco entre l\'inguas.

O m\'etodo usado pela glotocronologia moderna, 
desenvolvido por Swadesh na d\'ecada de 
1950, mede dist\^{a}ncias entre l\'inguas a partir 
do percentual de palavras com origem 
hist\'orica comum em uma lista. O ponto fraco desse 
m\'etodo \'e o grau de subjetividade 
presente no julgamento da dist\^{a}ncia.

Recentemente propusemos um m\'etodo automatizado 
que evita a subjetividade, 
cujos resultados podem ser replicados por estudos 
que usem a mesma base de dados e 
que n\~{a}o necessita nenhum conhecimento 
lingu\'istico espec\'ifico por parte do pesquisador. 
Al\'em do mais, o m\'etodo permite uma compara\c{c}\~{a}o 
r\'apida de um grande número de l\'inguas.

Aplicamos nosso m\'etodo aos grupos Indoeuropeu e 
Austron\'esio considerando, em cada caso, 
cinquenta l\'inguas diferentes. As \'arvores geneal\'ogicas 
resultantes s\~{a}o semelhantes 
\`as de estudos anteriores, mas com algumas diferen\c{c}as 
importantes na posi\c{c}\~{a}o de 
poucas l\'inguas e subgrupos. Acreditamos que essas 
diferen\c{c}as carregam informa\c{c}\~{o}es 
novas sobre a estrutura da \'arvore e sobre as rela\c{c}\~{o}es 
filogen\'eticas dentro das fam\'ilias.

\bigskip

\bigskip
\bigskip

Languages ​​evolve over time in a process in 
which reproduction, mutation and extinction
are all possible, similar to what happens to living organisms.
Using this similarity it is possible, in principle, to build family trees 
which show the degree of relatedness between languages.

The method used by modern glottochronology, developed by Swadesh in the 1950s, 
measures distances from the percentage of words with a
common historical origin. The weak point of this method is that subjective judgment
plays a relevant role.

Recently we proposed an automated method that avoids the subjectivity,
whose results can be replicated by studies that use the same database and
that doesn't require a specific linguistic knowledge.
Moreover, the method allows a quick comparison of a large number of languages.

We applied our method to the Indo-European and Austronesian families,
considering in both cases, fifty different languages. 
The resulting trees are similar to those of previous studies, but with 
some important differences in the position of few languages ​​and subgroups.
We believe that these differences carry new information on the structure 
of the tree and on the phylogenetic relationships within families.
\end{abstract}

\newpage

\section{Introduction}

Glottochronology tries to estimate the time at which languages diverged
with the implicit assumption that vocabularies change at a constant rate.
The idea, initially proposed by M. Swadesh \cite{Sw}, is to restrict the 
comparison to a list of terms which are common to all cultures and which 
concern the basic activities of humans.
The choice is motivated by the fact that these terms are learned during
childhood and they change very slowly over time.

The use of Swadesh lists in glottochronology 
is popular since half a century.
Glottochronologists use the percentage of shared {\it cognates}
in order to compute the distances between pairs of languages.
Divergence times are assumed to be, on average, logarithmically proportional 
to these {\it lexical} distances.
A recent example of the use of Swadesh 
lists and cognates to construct language 
trees are the studies of Gray and Atkinson 
\cite{GA} and Gray and Jordan \cite{GJ}.

Cognates are words inferred to have a common historical origin, 
their identification is often a matter of 
sensibility and personal knowledge. 
In fact, the task of counting the number of 
cognate words in the list is far from
trivial because cognates do not necessarily look similar.
Therefore, subjectivity plays a relevant role.
Furthermore,  results are often biased 
since it is easier for European or American scholars to find
out those cognates belonging to western languages.
For instance, the Spanish word {\it leche} and the Greek 
word {\it gala} are cognates.  In fact, {\it leche} comes
from the  Latin {\it lac} with genitive form {\it lactis},
while the genitive form of {\it gala} is {\it galactos}.
Also the English {\it wheel} and Hindi {\it cakra} are cognates. 
These two identifications are possible
because of our historical records, hardly they would 
have been possible for languages, let's say, of Central Africa
or Australia. 

The idea of measuring relationships among languages
using vocabulary, seems to have its roots in the work of 
the French explorer Dumont D'Urville.
He collected comparative words lists of various
languages during his voyages aboard the Astrolabe 
from 1826 to 1829 and, in his work about 
the geographical division of the Pacific \cite{Urv},
he proposed a method to measure the degree of relation 
among languages.
He used a core vocabulary of 115 base terms
which, impressively, contains all but three of 
the terms in Swadesh's 100-item list.
Then, he assigned a distance from 0 to 1 to any pair 
of words with the same meaning 
and finally he was able to determine the degree of 
kinship for any pair of languages.

In our work we used an automated method
which avoids subjectivity
so that our results can be replicated by other scholars assumed that
the database used is the same.
For any language we wrote down a list of the same 200 words according
to the original choice of Swadesh \cite{Sw}, then 
we compared words with same meaning belonging to
different languages only considering orthographical differences.
This may appear reductive since words may look similar by chance,
while cognate words may have a completely different orthography,
but we will try to convince the reader that indeed this is a simpler,
more objective and more efficient choice with respect
to the traditional glottochronological approach.

To be precise, we defined the distance between two languages 
(section two) by computing a normalized Levenshtein distance among words 
with the same meaning and by averaging on the two 
hundred terms contained in the lists \cite{footnote}.
The normalization, which takes into account the word's length,
plays a crucial role, and no sensible results would have been found 
without it. We applied this strategy to the Indo-European and 
the Austronesian families considering, in both cases, fifty different 
languages and obtaining two matrices of distances with 1225
non trivial entries.

These distances can be transformed, by a simple logarithmic rule,
in separation times (section three) and  two genealogical trees can be
generated (section four) using the Unweighted Pair Group Method 
Average (UPGMA) \cite{UPGMA}.
The trees are similar to those found by 
\cite{GA} and \cite{GJ} with some important differences concerning
the position of few languages and subgroups.
Indeed, we think that these differences carry some new
information about the structure of the tree and about
the position of some languages as Malagasy and Romani.

\section{Definition of lexical distance}

We start by our definition of lexical distance between two words,
which is a variant of the Levenshtein distance. 
The Levenshtein distance is simply the minimum
number of  insertions, deletions, or substitutions of a
single character needed to transform one word into the other.
Our definition is taken as the Levenshtein distance divided 
by the number of characters of the longer of the two.

More precisely, given two words $\alpha_i$ and $\beta_j$
(the Greek letter indicates the language while
the Latin subscript indicates the meaning)
their distance $d(\alpha_i, \beta_j)$ is given by

\begin{equation}
d(\alpha_i, \beta_j)= 
\frac{d_l(\alpha_i, \beta_j)}{l(\alpha_i, \beta_j)}
\label{wd}
\end{equation}
where $d_l(\alpha_i, \beta_j)$ is the 
Levenshtein distance between the two words
and $l(\alpha_i, \beta_j)$ is the
number of characters of the longer of the two.
Therefore, the distance can take any value between 0
and 1 and, obviously, $d(\alpha_i, \alpha_i)=0$.

The reason why we renormalize can be understood from
the following example.
Consider the case in which a single substitution 
transforms one word into another
with the same length.  
If they are short, let's say 2 characters, 
they are very different. 
On the contrary,
if they are long, let's say 8 characters, it is reasonable to say
they are very similar. 
Without renormalization, their distance would be 
the same and equal 1, regardless of their length. 
Instead, introducing the normalization factor, in the first case 
the distance is $\frac{1}{2}$,
whereas in the second, it is much smaller
and equal to $\frac{1}{8}$. 

For any language, the first step is to write down a list
of the words corresponding to the Swadesh's choice of meanings. 
Then, the lexical distance between a pair of languages 
is defined as the average of the distance between all pair of words
corresponding to the same meaning.  
Assume that the number of languages is $N$
and the list of words for any language contains
$M=200$ items. 
Any language in the family is labeled a Greek letter
(say $\alpha$) and any word of that language by 
$\alpha_i$ with $1 \leq i \leq M$. Then, two words $\alpha_i$ and 
$\beta_j$  in the languages $\alpha$ and $\beta$ 
have the same meaning if $i=j$.

The above defined distance between two languages is written symbolically as 

\begin{equation}
D(\alpha, \beta)=  \frac{1}{M} \sum_{i=1}^M
d(\alpha_i, \beta_i)
\label{ld}
\end{equation}
Notice that only pairs of words with the 
same meaning are 
used in this definition. It can be seen that $D(\alpha, \beta)$
is always in the interval [0,1] and, obviously, $D(\alpha, \alpha)=0$. 

The database used here \cite{footnote} 
to construct the phylogenetic tree is 
composed by $N=50$ languages of the Indo-European family 
and $N=50$ languages of the Austronesian one. 
The main source for the Indo-European database is the file prepared by 
Dyen et al. in \cite{D} which contains the 
Swadesh list of 200 words for 96 languages.   
Many words are missing in \cite{D}  but for our choice of 50 languages
we have filled most of the gaps and corrected some errors
by finding the words on dictionaries freely available on the web.  
For the Austronesian family we used as the main source the 
lists contained in the huge database \cite{NZ}.
The lists in \cite{NZ} contain more than 200 words but the meanings
do not coincide completely with those of the original Swadesh
list \cite{Sw}.
For our 50 Austronesian languages 
we have retained only those words corresponding to the
meanings which are also in the original Swadesh list.
There are many gaps due to this incomplete overlap 
and because of many missing words in \cite{NZ}.
Also in this case we have filled some of the gaps 
by finding the words on the web
and, in the case of Malagasy, by direct knowledge of the language.

For some of the languages in our lists \cite{footnote}
there are still few missing words. 
When a language has one or more missing words, 
these are simply not considered in the average
that gives the lexical distance between two languages. 
This implies that for some pairs of languages,  
the number of compared words is not 200, but smaller. 
There is no bias in this procedure, the only effect
is that the statistic is slightly reduced.  
Indeed, the definition (\ref{ld}) is modified,
by replacing $M=200$ with the number of
word pairs with same meaning existing in both lists
and the sum goes on all these pairs.  

In the database only the English alphabet is used (26 characters plus
space); those languages written in a different alphabet 
(i.e. Greek etc.) were already transliterated into the English
one in \cite{D}.
Furthermore, in \cite{NZ} many additional characters are used 
which we have eliminated so that also in this case
we reduce to the English alphabet plus space. 
Our database is available at \cite{footnote}.

The result of the analysis described above are two  
$50  \times  50$ upper triangular matrices
whose entries are the 1225 non-trivial 
lexical 
distances $D(\alpha, \beta)$ between all pairs
in a family.
Indeed, our method for computing distances is a very simple operation, 
that does not need any specific linguistic 
knowledge and it requires a minimum of computing time.

\section{Time distance between languages}

A phylogenetic tree can already be built from one of these matrices,
but this would only give the topology of the tree, 
whereas the absolute time scale would be missing.
In order to have this quantitative information, some hypotheses  
on the time evolution of lexical distances are necessary.
We assume that the lexical distance among words, on one side tends
to grow due to random mutations and on the other side may decrease 
since different words may become more similar by accident or, 
more likely, by language borrowings.

Therefore, the  distance $D$ between two given languages can be thought to  
evolve according to the simple differential equation
\begin{equation}
\label{diffeq}
\dot{D}=a \,(1-D) -b D
\end{equation}
where $\dot{D}$ is the time derivative of $D$.
The positive parameter $a$ is related to the increasing of $D$ 
due to random permutations, deletions or substitutions of characters 
(random mutations) while the positive parameter $b$ considers
the possibility that two words become more similar 
by a ``lucky'' random mutation or by words borrowing
from one language to the other or both from a third one. 
Since $a$ and $b$ are constant, it is implicitly
assumed that mutations and borrowings occur at a constant rate.

Note that with this choice, word substitution is statistically equivalent 
to the substitution of all characters in the word itself.
The first reason for this approximation is reducing the number 
of parameters in the model. The second, and more important, 
is that it is very hard to establish if a word has changed because many 
characters have been replaced or if the whole word has been replaced. 
Only historical records would give this information but
this would imply again a subjective analysis that we want to avoid 
within our model. 

At time $T=0$ two languages begin
to separate and the lexical distance $D$ is zero.
With this initial condition the above equation can be solved
and the solution can be inverted. The result
is a relation which gives the separation time $T(\alpha, \beta)$ 
between two languages $\alpha$ and $\beta$
in terms of their lexical distance $D(\alpha, \beta)$
\begin{equation}
T(\alpha, \beta)= -\epsilon \,\ln(1 - \gamma D(\alpha, \beta))
\label{time}
\end{equation}
The values for the parameters $\epsilon = 1/(a + b)$
and  $\gamma = (a+b )/a$ can be fixed 
experimentally by considering  two pairs of 
languages whose separation time (time distance) is known. 
We have chosen a distance of 1600 years between Italian and French
and a distance of 1100 years between Icelandic and Norwegian. 
The resulting values of the parameters 
are $\epsilon =1750$ and
$\gamma=1.09$,  which correspond to the values $a \cong 5*10^{-4}$ 
and $b \cong 6*10^{-5}$.
This means that similar words may become more different at a rate
that is about ten times the rate at which different words 
may become more similar. It should be noticed that (\ref{time})
closely resembles the fundamental formula of glottochronology.
We use this choice of the parameters
both for the Indo-European and Austronesian families.

A time distance $T(\alpha, \beta)$ is then computed for all pairs of languages 
in the database, obtaining two $50 \times  50 $ upper triangular 
matrices with $1225$ non-trivial entries. 
These matrices preserve the topology of the lexical distance matrices
but they contain all the information concerning absolute time scales.

\section{Trees}

\begin{figure}
\epsfysize=20.0truecm \epsfxsize=18.0truecm
\centerline{\epsffile{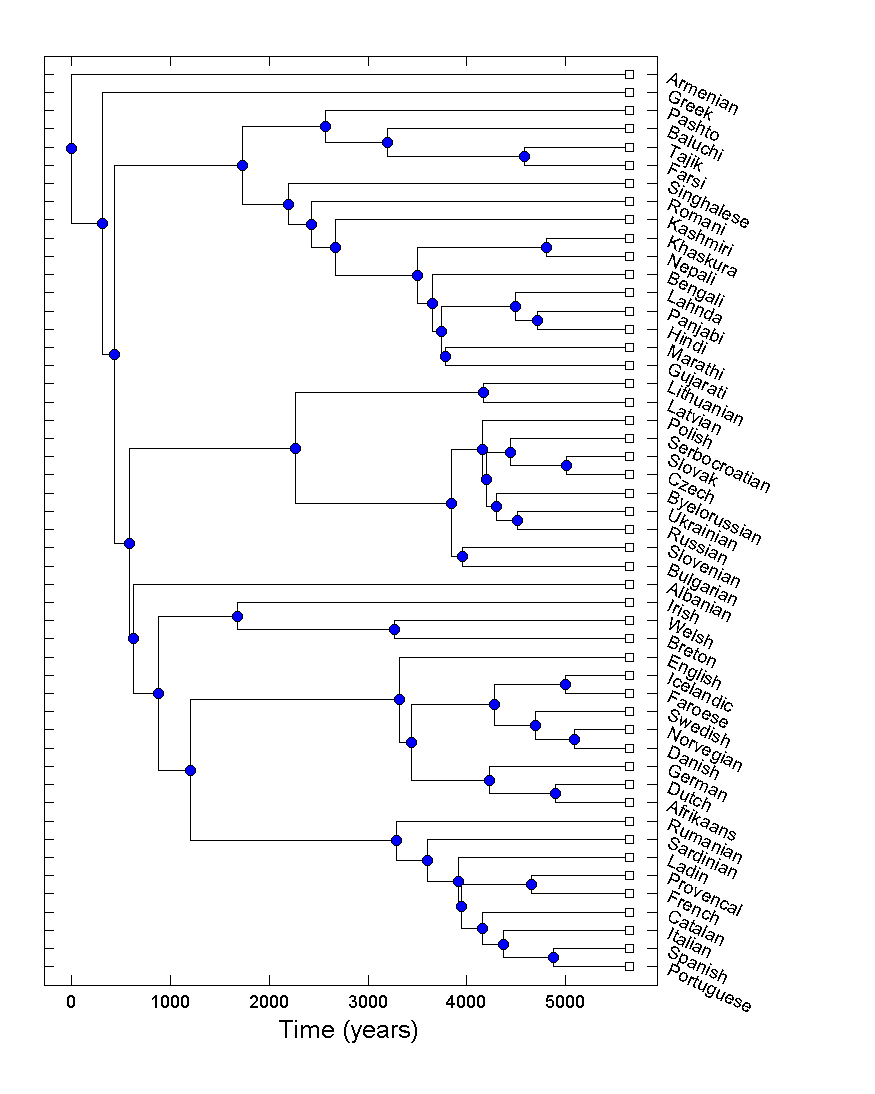}} 
\caption{Indo-European phylogenetic tree constructed from the matrix
of distances using UPGMA.}
\label{fig1}
\end{figure}

\begin{figure}
\epsfysize=20.0truecm \epsfxsize=18.0truecm
\centerline{\epsffile{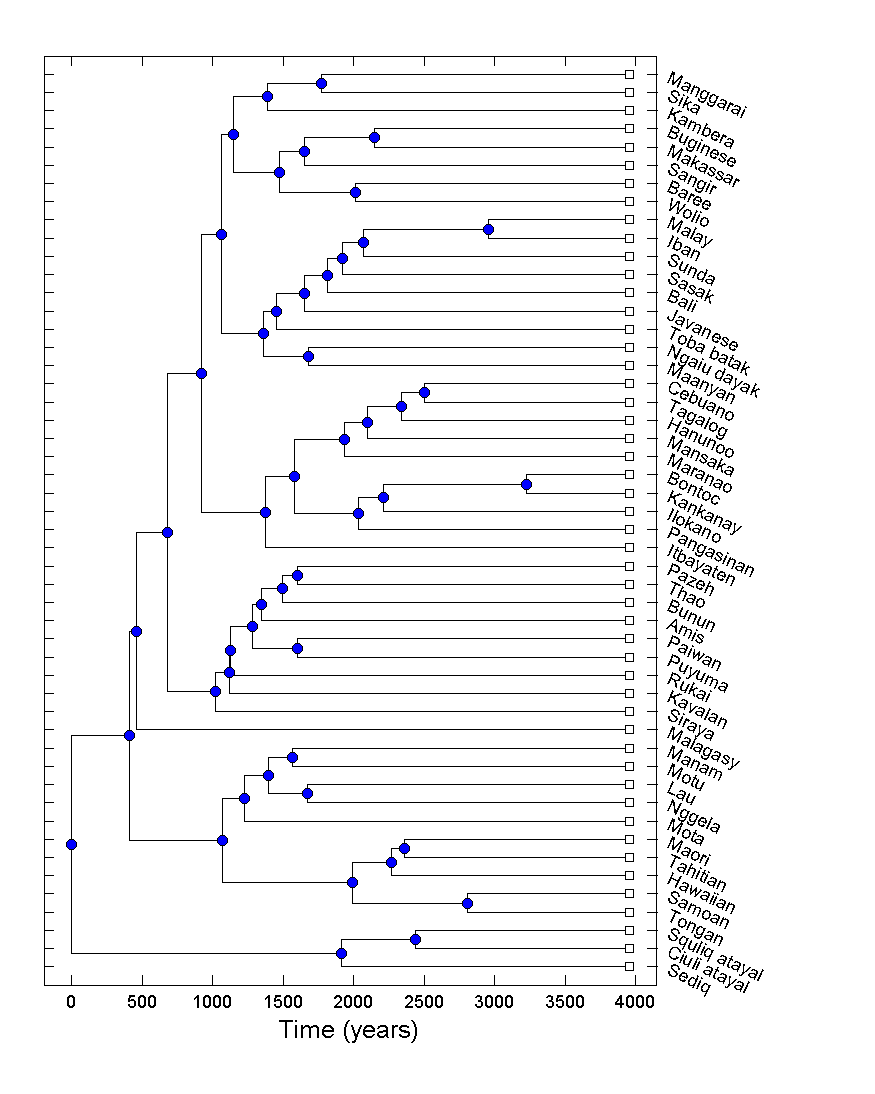}} 
\caption{Austronesian phylogenetic tree constructed from the matrix
of distances using UPGMA.}
\label{fig2}
\end{figure}

Phylogenetic trees in Fig. \ref{fig1} and in Fig. \ref{fig2} 
are constructed from the matrix using the Unweighted Pair Group 
Method Average (UPGMA) \cite{UPGMA}.
We use UPGMA for its coherence with the coalescence process of Kingman 
type \cite{K}.
In fact, the process of languages separation and extinction closely 
resembles the population dynamics associated with haploid reproduction 
which holds for simple organisms 
or for the mitochondrial DNA of complex ones.
This dynamics, introduced by Kingman, has been extensively
studied and described, see for example \cite{S,SD}.
It should be considered that in the model of Kingman,
time distances have the objective meaning of measuring time
from separation while in our realistic case the
time distances are reconstructed from lexical distances.
In this reconstruction we assume  that lexical mutations and
borrowings happen at a constant rate.
This is true only on average, since there is an 
inherent randomness in this process \cite{PRS}
which is not taken into account by
the deterministic differential equation (\ref{diffeq}).
Furthermore, parameters $a$ and $b$  may vary from a pair of languages
to another and also they may vary in time 
according to historical conditions.

To check the stability of the phylogenetic trees we computed many trees 
in which some languages were removed randomly. The computation of these 
trees shows a strong stability in the main features of the trees, 
namely, all the large branches remain the same if some
of their leaves are removed. 

The Indo-European tree in Fig. \ref{fig1} is similar to the one 
in \cite{GA} but there are some important differences.
First of all, the first separation concerns Armenian, which is 
a isolated branch close to the root, while the other branch
contains all the remaining Indo-European languages.
Then, the second separation is that of Greek,
and only after there is a separation between the European branch and the
Indoiranian one.
This is at variance with the tree in \cite{GA}, 
since therein the separation at the root
gives origin to two branches, one with Indoiranian languages 
plus Armenian and Greek, the other with European languages.
The position of Albanian is also different: in our case it is
linked to European languages while in \cite{GA} it goes with Indoiranian ones. 

Finally, in Fig. \ref{fig1} the Romani language is correctly located together 
with the Indian languages but it is not as close to Singhalese 
as reported in \cite{GA}.
Romani is the language of Roma and Sinti,
it turns out that the closest three languages are Nepali, Bengali 
and Khaskura. 
Nepali and Kaskura are spoken in Nepal and 
northern India (the second is the language of Gurkhas), while
Bengali is spoken in Northeast India and Bangladesh.
This implies a geographical origin in Northern India for 
Roma's and Sinti's,  according to the beliefs 
of the majority of researchers.
Our results are different from those found in \cite{GA}
where a close relationship Romani/Singhalese is detected.

Also the tree in Fig. \ref{fig2} is similar to the one in \cite{GJ} 
but differences here are more important.
The first separation concerns the Atayal Formosan languages 
in a branch, while the Paiwan Formosan languages 
are in the other branch together with all the 
Malayo-Polynesian languages.
This result, if confirmed, would suggest two different waves 
of migration from Formosa. An alternative, and more likely,
explanation could be an Austronesian homeland outside Taiwan \cite{Bl}.  

Finally, the Malagasy language is isolated
even if the closest language is Maanyan,  
which belongs to the south-east Barito group of languages 
spoken in Kalimantan \cite{Dahl}. 
Surprising, the second closest language is Maranao 
which is spoken in Philippines 
while the third closest is Buginese spoken in south Sulawesi.
The fact that Malagasy, as expected, is very close to Maanyan,
but other close languages are not in Kalimantan
could suggest a multiple origin.
It should be mentioned that Malagasy has many loanwords 
from Malay, particularly in the domain of maritime life and 
navigation. 
 
The main problem is that it is unlikely that Maanyan Dayaks 
undertook the spectacular migrations from Kalimantan to Madagascar, since
they are forest dwellers with river navigation skills only.
A possible explanation is that they were brought there as slaves by Malay 
seafarers, which also took slaves from other parts of Southeast Asia.
If the south-east Barito speakers formed the majority
in the initial group,  their language could have constituted the core 
element of what later became Malagasy. 
In this way Malagasy absorbed words of the Austronesian
(and African) languages of the other slaves and of the Malay seafarers.

\section{Discussion and conclusions}

Automated language classification is very 
important for the understanding of language phylogeny. 
In particular, it is very useful for those languages 
for which finding cognacy relations is a difficult task. 
It also permits to classify a huge number of languages in a 
very short time by using computer programs.
 
The automated method described here was later used and developed 
by another large team of scholars that placed the method at 
the core of an ambitious project, the ASJP (The Automated Similarity 
Judgment Program) whose aim, in the words of its proponents, is
"..achieving a computerized 
lexicostatistical analysis of ideally all the world's languages" \cite{Wic}. 

In \cite{SP3},  we have completed our research
by a careful study of the words stability problem.
This study allows us to find the optimal length of the lists of words 
to be used for the phylogeny reconstruction of a family of languages. 
The method is also automatic and gives lists of stable words 
which depend upon the language family,  
according to its specific cultural traits. 

More recently, together with other scholars \cite{Bl},
we have used lists of automatically computed distances  
for a deeper analysis of the relationships among languages. 
The point is that a tree is only an approximation,
which skips complex phenomena as horizontal transfer.
Our method, which gives a geometric representation,
correctly finds out language clusters but also gives 
a lot of new information. 
It allows, for example, a more accurate understanding
of some important topics, as migration patterns 
and homeland locations of the families of languages. 

Finally our method was able to resolve some of the mysteries
concerning the settlement of Madagascar \cite{SPVW,S2}.

\section*{Acknowledgments}

The ideas and method presented here are mostly
the fruit of joint work with Filippo Petroni, while
recent developments have been obtained in collaboration with
Philippe Blanchard and Dimitri Volchenkov and S\o{}ren Wichmann.
We warmly thank Armando G. M. Neves for helpful discussions
and for a critical reading of an early version of this paper.
We are indebted with Lydie Irene Andriamiseza for improvements 
in the Malagasy database and related Austronesian Swadesh lists. 
Finally we thank S.J. Greenhill, R. Blust and R.D. Gray,
for the authorization to use their 
{\it Austronesian Basic Vocabulary Database} \cite{NZ}.

\newpage

\newpage

\end{document}